\documentclass[runningheads]{llncs}
\usepackage[T1]{fontenc}
\usepackage{amsmath,booktabs,amssymb}
\usepackage{changepage}
\usepackage{multirow}
\usepackage{subfig}
\usepackage{algcompatible}
\usepackage{amsmath}
\usepackage{adjustbox}
\usepackage[table]{xcolor}
\usepackage{dsfont}
\usepackage{booktabs}
\usepackage{multirow}
\usepackage{graphicx}
\usepackage[%
    colorlinks=true,
    pdfborder={0 0 0},
    linkcolor=blue
]{hyperref}
\hypersetup{
    colorlinks=true,
    linkcolor=blue,
    filecolor=magenta,      
    urlcolor=blue,
    citecolor=red,
}
\usepackage{paralist}
\usepackage{soul}
\usepackage{scalerel}
\usepackage{tikz}
\usetikzlibrary{svg.path}

\usepackage{graphicx}

\begin{document}
\title{Addressing Class Imbalance in Semi-supervised Image Segmentation: A Study on Cardiac MRI}

\titlerunning{Class Imbalance in Semi-supervised Segmentation}

\author{Hritam Basak\inst{1} \thanks{Corresponding author} \and
Sagnik Ghosal\inst{1} \and
Ram Sarkar\inst{2}}
\authorrunning{H. Basak et al.}

\institute{Dept. of Electrical Engineering, Jadavpur University, Kolkata, India \and
Dept. of Computer Science and Engineering, Jadavpur University, Kolkata, India\\
\email{\{hritambasak48, sagnikghosal1999, ramjucse\}@gmail.com}}

\maketitle              
\begin{abstract}
Due to the imbalanced and limited data, semi-supervised medical image segmentation methods often fail to produce superior performance for some specific tailed classes. Inadequate training for those particular classes could introduce more noise to the generated pseudo labels, affecting overall learning. To alleviate this shortcoming and identify the under-performing classes, we propose maintaining a confidence array that records class-wise performance during training. A fuzzy fusion of these confidence scores is proposed to adaptively prioritize individual confidence metrics in every sample rather than traditional ensemble approaches, where a set of predefined fixed weights are assigned for all the test cases. Further, we introduce a robust class-wise sampling method and dynamic stabilization for a better training strategy. Our proposed method considers all the under-performing classes with dynamic weighting and tries to remove most of the noises during training. Upon evaluation on two cardiac MRI datasets, ACDC and MMWHS, our proposed method shows effectiveness and generalizability and outperforms several state-of-the-art methods found in the literature.  

\end{abstract}

\section{Introduction}
\label{sec:Introduction}
Recent years have witnessed a significant improvement in medical image analysis using deep learning tools \cite{basak2020comparative,basak2021dfenet,basak2020f,basak2021cervical}. In the context of medical imaging, access to large volumes of labelled data is difficult owing to the high cost, required domain-specific expertise, and protracted process involved in generating accurate annotations \cite{basak2022mfsnet,basak2022exceedingly}. Using a lesser amount of training data, on the other hand, significantly affects the model's performance. To solve this bottleneck, researchers shifted towards the domain of Semi-Supervised Learning (SSL) which exploits unlabeled data information to compensate for the substantial requirement of data annotation \cite{chen2019self}. 
Recently SSL-based medical image segmentation strategies have been widely adopted due to their competing performance and ability to learn from very few annotations. To this end, adversarial learning is a very promising direction where Peng et al. \cite{peng2020deep} proposed an adversarial co-training strategy to enforce diversity across multiple models. Li et al. \cite{li2021semi} took a generative adversarial-based approach for cardiac magnetic resonance imaging (MRI) segmentation, where they proposed utilizing the predictive result as a latent variable to estimate the distribution of the latent. Nie et al. \cite{nie2018asdnet} proposed ASDNet, an adversarial attention-based SSL method utilizing a fully convolutional confidence map. Besides, Luo et al. \cite{luo2021semi} proposed a dual-task consistent network to utilize geometry-aware level-set as well as pixel-level predictions.   
Contrastive Learning (CL) based strategies \cite{chaitanya2020contrastive,zeng2021positional} have also been instrumental in this purpose by enforcing representations in latent space to be similar for similar representations. Chaitanya et al. \cite{chaitanya2020contrastive} showed the effectiveness of global and local contexts to be of utmost importance in contrastive pretraining to mine important latent representations. Lately, Peng et al. \cite{peng2021self} tried to address the limitations of these CL-based strategies by proposing a dynamic strategy in CL by adaptively prioritizing individual samples in an unsupervised loss. Other directions involve consistency regularization \cite{xie2021intra}, domain adaptation \cite{xia2020uncertainty}, uncertainty estimation \cite{wang2021tripled}, etc.


However, a significant problem to train the existing SSL-based techniques is that they are based on the assumption that every class has an almost equal number of instances \cite{stanescu2014semi}. On the contrary, most real-life medical datasets have some classes with notably higher instances in training samples than others, which is technically termed as $class$ $imbalance$. For example, the class $myocardium$ in ACDC \cite{bernard2018deep} is often missing or too small to be detected in apical slices, leading to substandard segmentation performance for this class \cite{bernard2018deep}.  This class-wise bias affects the performance of traditional deep learning networks in terms of convergence during the training phase, and generalization on the test set \cite{japkowicz2002class}.

This paper addresses a relatively new research topic, called the class imbalance problem in SSL-based medical image segmentation. Though explicitly not focusing on class imbalance, our proposed method aims to improvise the segmentation performance of tail classes by keeping track of category-wise confidence scores during training. Furthermore, we incorporate fuzzy adaptive fusion using the Gompertz function where priority is given to individual confidence scores in every sample in this method rather than traditional ensemble approaches (average, weighted average, etc.), where a set of predefined fixed weights is assigned for all the test cases.

\section{Proposed Method}

\begin{figure}[tbp]
    \centering
    \includegraphics[width=0.85\columnwidth]{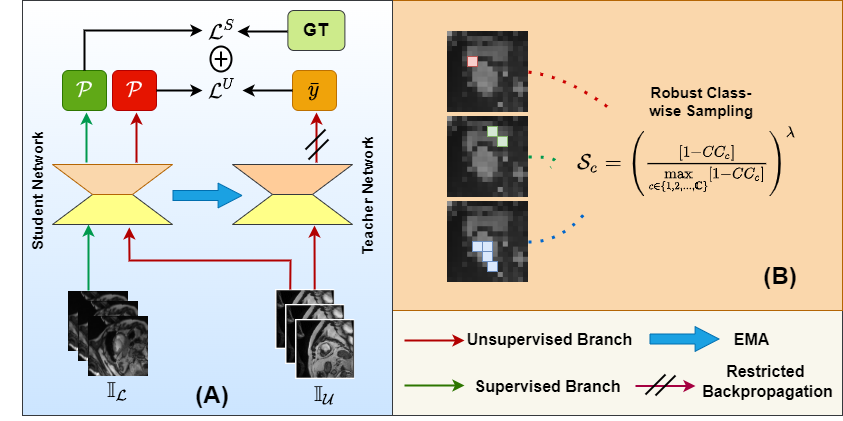}
    \caption{Overall training strategy of the proposed method: (A) the basic student-teacher network, used as a backbone in our work, and (B) the Robust Class-wise Sampling that adaptively samples more pixels from the under-performing classes}
    \label{overall_figure}
\end{figure}

Let us assume that the dataset consists of $\mathbb{N}_1$ number of labelled images $\mathbb{I}_\mathcal{L}$ and $\mathbb{N}_2$ number of unlabelled images $\mathbb{I}_\mathcal{U}$ (where $\{\mathbb{I}_\mathcal{L}, \mathbb{I}_\mathcal{U}\} \in \mathbb{I}$ and $\mathbb{N}_1<<\mathbb{N}_2$). First, we define the standard student-teacher architecture, and then formulate our dynamic training strategy by redefining the loss terms. 

\subsection{Basic Student-Teacher Framework}
Similar to \cite{tarvainen2017mean}, our network consists of a student-teacher framework, where the network learns through the student branch only, and the weights of the teacher model are updated by using an Exponential Moving Average (EMA). The student model is used to generate pseudo labels $\bar{\mathcal{Y}}$ on both labelled data $\mathbb{I}_\mathcal{L}$ with weak augmentations and unlabelled data $\mathbb{I}_\mathcal{U}$ with strong augmentations. In contrast, the teacher model only generates pseudo labels on weakly augmented unlabelled data $\mathbb{I}_\mathcal{U}$. The base student-teacher model is shown in \autoref{overall_figure}\textcolor{blue}{(A)}. The standard supervised and unsupervised loss functions ($\mathcal{L}^S$ and $\mathcal{L}^U$) can, therefore, be defined as:
\begin{equation}
    \mathcal{L}^S = \frac{1}{\mathbb{N}_1}\sum\limits_{n=1}^{\mathbb{N}_1}   \frac{1}{h \times w} \sum\limits_{i=1}^{h\times w} \mathcal{L}_{CE}(\mathcal{P}_{n,i}, GT_{n,i})
\end{equation}

\begin{equation}\label{unsupervised_loss}
    \mathcal{L}^U = \frac{1}{\mathbb{N}_2}\sum\limits_{n=1}^{\mathbb{N}_2}   \frac{1}{h \times w} \sum\limits_{i=1}^{h\times w} \mathcal{L}_{CE}(\mathcal{P}_{n,i}, \bar{{y}}_{n,i})
\end{equation}
where, $h\times w$ is the image dimension, $\mathcal{L}_{CE}$ represents the standard pixelwise cross-entropy loss, $\mathcal{P}_{n,i}$ represents the prediction of $i^{th}$ pixel of $n^{th}$ image, $GT$ is the ground truth label, and $\bar{y}_{n,i}\in{\bar{\mathcal{Y}}}$ is the generated pseudo label for the corresponding pixel of the corresponding image.

\subsection{Dynamic Class-aware Learning}

\subsubsection{Formulation of Confidence Array}\label{confidence-array}
Some of the methods in literature \cite{li2020overcoming,hu2021adaptive} address the imbalanced training in natural scene images by relying on class-wise sample counts followed by ad-hoc weighting and sampling strategies. 
To this end, we propose maintaining three different performance indicators, namely $Entropy$, $Variance$, and $Confidence$, in a class-wise confidence array to assess the performance of every class. We define $Entropy$ indicator $\mathbb{E}_c$ for class $c$ as:
\begin{equation}
    \mathbb{E}_c = \frac{1}{\mathbb{N}_1}\sum\limits_{n=1}^{\mathbb{N}_1} \frac{1}{\mathbb{N}_c^n}\sum\limits_{i=1}^{\mathbb{N}_c^n}\sum\limits_{j=1}^\mathbb{C} \mathcal{P}_{n,i}^{c}\log \mathcal{P}_{n,i}^{j}\:; \:\: \forall c\in\{1,2,3,...,\mathbb{C}\}
\end{equation}
where $\mathcal{P}_{n,i}^j$ is the $j^{th}$ channel prediction for the $i^{th}$ pixel of $n^{th}$ image. Similarly, we define $Variance$ and $Confidence$ indicators $\mathbb{V}_c$ and $\mathbb{C}on_c$ respectively as:
\begin{equation}
\begin{split}
    \mathbb{V}_c = \frac{1}{\mathbb{N}_1}\sum\limits_{n=1}^{\mathbb{N}_1} \frac{1}{\mathbb{N}_c^n}\sum\limits_{i=1}^{\mathbb{N}_c^n}\left( \max\limits_{j\in \{1,2,..,\mathbb{C} \}}[\mathcal{P}_{n,i}^{j}] - \mathcal{P}_{n,i}^{c}  \right);\:\:
    \forall c\in\{1,2,3,...,\mathbb{C}\}
\end{split}
\end{equation}

\begin{equation}
    \mathbb{C}on_c = \frac{1}{\mathbb{N}_1}\sum\limits_{n=1}^{\mathbb{N}_1} \frac{1}{\mathbb{N}_c^n}\sum\limits_{i=1}^{\mathbb{N}_c^n} \mathcal{P}_{n,i}^{c} \:; \:\: \forall c\in\{1,2,3,...,\mathbb{C}\}
\end{equation}

\subsubsection{Fuzzy Confidence Fusion}\label{fuzzy_fusion}
We combine the three class-wise performance indicators $\mathbb{E}_c$, $\mathbb{V}_c$, and $\mathbb{C}on_c$ to generate the final $CC$ score using a fuzzy fusion scheme. Gompertz function was experimentally adopted for fuzzy fusion as explained in the supplementary material. 
First, we generate a class-wise fuzzy rank of different performance indicators using a re-parameterized Gompertz function \cite{kundu2021fuzzy} as:
\begin{equation}
    \mathbb{R}_c^k = 1-e^{-e^{-2\cdot norm (x_c^k)}}, \:\: \text{where }x_c^k\in\{\mathbb{E}_c, \mathbb{V}_c, \mathbb{C}on_c\},
\end{equation}
where, $norm()$ signifies normalization function, $\mathbb{R}_c^k$ is in range $[0.127,0.632]$, where a higher confidence score gives better (lower) rank. The selection of the objective fuzzy function is based on model performance. For detailed analysis, please refer to the supplementary file. Now, if $M^k$ represents top $m$ ranks for class $c$, then we compute a complement of confidence factor sum ($CCF_c$) and fuzzy rank sum ($FR_c$) as follows:
\begin{equation}
    CCF_c = \sum_{k} \begin{cases}
            norm(x_c^k), &\text{if } \mathbb{R}_c^k \in M^k \\
            P_{c}^{CCF}, &\text{otherwise }
        \end{cases}
        \:\text{and}\:\:
         FR_c = \sum_k \begin{cases}
    \mathbb{R}_c^k, &\text{if } \mathbb{R}_c^k \in M^k \\
    P_c^{FR}, &\text{otherwise }
    \end{cases}
\end{equation}
where, $P_{c}^{CCF}$ and $P_c^{FR}$ are the penalty values (set to 0 and 0.632 respectively) for class $c$ to suppress the unlikely winner. The final cumulative confidence ($CC$) score is thereafter computed as:
\begin{equation}\label{fuzzy_equation}
    CC_c = CCF_c \times FR_c \:\:\forall c\in\{1,2,...\mathbb{C}\}
\end{equation}
The obtained $CC$ score is updated after $t^{\text{th}}$ training iteration as:
\begin{equation}
    CC_c^{t} \longleftarrow \alpha CC_c^{t-1}+(1-\alpha) CC_c^{t} \: ;\\
    \forall c\in\{1,2,3,...,\mathbb{C}\}
\end{equation}
where, $\alpha$ is the momentum parameter, set to $0.999$ experimentally.

\subsubsection{Robust Class-wise Sampling}
We obtain the category-wise confidence score and identify the under-performing classes as described in \autoref{confidence-array}. To alleviate the problem of class-wise training bias, i.e., preventing well-performing classes from overwhelming model training and anchoring the training on a sparse set of under-performing classes, we propose a class-wise sampling rate $\mathcal{S}_c$ as:
\begin{equation}\label{s_c}
    \mathcal{S}_c = \left( \frac{[1-CC_c]}{\max\limits_{c\in\{1,2,...,\mathbb{C}\}} [1-CC_c]}  \right)^\lambda
\end{equation}
where $\lambda$ is a tunable parameter. Instead of sampling all the available pixels for the unsupervised loss formulation, we sample random pixels from class $c$ with sampling rate of $\mathcal{S}_c$. So $\mathcal{L}^U$ in \autoref{unsupervised_loss} can be reformulated as:
\begin{equation}\label{sampling-loss}
    \mathcal{L}^U = \frac{1}{\mathbb{N}_2}\sum\limits_{n=1}^{\mathbb{N}_2}   \frac{1}{\left(\sum\limits_{i=1}^{h\times w}\mathds{1}_{n,i}\right)} \sum\limits_{i=1}^{h\times w} \mathds{1}_{n,i}\mathcal{L}_{CE}(\mathcal{P}_{n,i}, \bar{{y}}_{n,i})
\end{equation}
where $\mathds{1}$ is the binary value operator. The value $\mathds{1}_{n,i}=0$ if the $i^{th}$ pixel from the $n^{th}$ image is not sampled according to sampling rate $\mathcal{S}_c$, otherwise set to $1$. \autoref{overall_figure}\textcolor{blue}{(B)} represents the proposed class-wise sampling strategy.

\subsubsection{Dynamic Training Stabilization}
As the model performance strictly relies upon the quality of pseudo labels, the under-performing categories insert a significant amount of noise in the pseudo label, hindering the training process. Methods in literature \cite{ke2020guided} set a higher threshold value to remove the under-performing classes, although this firm criterion leads to lower recall value for those categories, affecting the overall training. We utilize a dynamic modulation of weights to alleviate this problem for better training stabilization. This aims to redistribute the loss contribution from convincing and under-performing samples, i.e., more weights to the convincing classes. The unsupervised loss in \autoref{sampling-loss} can be reformulated as:
\begin{equation}
    \mathcal{L}^U = \frac{1}{\mathbb{N}_2}\sum\limits_{n=1}^{\mathbb{N}_2}   \frac{1}{\left(\sum\limits_{i=1}^{h\times w}  \mathcal{W}_{n,i}  \right)} \sum\limits_{i=1}^{h\times w} \mathcal{W}_{n,i}\mathcal{L}_{CE}(\mathcal{P}_{n,i}, \bar{{y}}_{n,i})
\end{equation}
where $\mathcal{W}_{n,i}$ is the weight provided for the $i^{th}$ pixel in $n^{th}$ image in the final unsupervised loss formulation, and can be defined as:
\begin{equation}\label{w_ni}
    \mathcal{W}_{n,i} = \mathds{1}_{n,i}\max\limits_{c\in\{1,2,...,\mathbb{C}\}}[\mathcal{P}_{n,i}]^\beta
\end{equation}
where $\beta$ is a tunable parameter. 
The final loss function is computed as:
\begin{equation}
    \mathcal{L}_{total} = \mathcal{L}^U+\zeta\mathcal{L}^S,
\end{equation}
where $\zeta$ is a tunable parameter. The value of $\zeta$ decreases with an increase in the number of iterations, limiting the contribution of supervised loss term $\mathcal{L}^S$ in the overall loss in the later stage of training.

\section{Experiments and Results}\label{results}
\subsection{Dataset and Implementation Details}
The model is evaluated on two publicly available cardiac MRI datasets. (1) the \textbf{ACDC dataset} \cite{bernard2018deep}, hosted in MICCAI17, contains 100 patients' cardiac MR volumes, where for every patient it has around 15 volumes covering the entire cardiac cycle, and expert annotations for left and right ventricles and myocardium. (2) The \textbf{MMWHS dataset} \cite{zhuang2016multi} consists of 20 cardiac MRI samples with expert annotations for seven structures: left and right ventricles, left and right atrium, pulmonary artery, myocardium, and aorta. The datasets are distributed into a $4:1$ ratio of training and validation sets for both cases. To validate the model performance on different label percentages, we use $1.25\%$, $2.5\%$, and $10\%$ labelled data from ACDC and $10\%$, $20\%$, and $40\%$ labelled data from MMWHS for training purposes (label percentage is taken in accordance to other methods in literature). The values of $\beta$ in \autoref{w_ni} and $\lambda$ in \autoref{s_c} are taken as $1.5$ and $2.5$, respectively (experimental analysis in supplementary file). The experimentation is implemented using Tesla K80 GPU with 16GB RAM. An SGD optimizer with an initial learning rate $1e-4$, momentum of $0.9$, and weight decay of $0.0001$ was employed. Three widely used metrics are used for the evaluation purpose: Dice Similarity Score (DSC), Average Symmetric Distance (ASD), and Hausdorff Distance (HD) \cite{basak2022embarrassingly}. 

\subsection{Quantitative Performance Evaluation}
\begin{figure}[tbp]
 \centering
    \subfloat[ACDC]{\includegraphics[scale = 0.4]{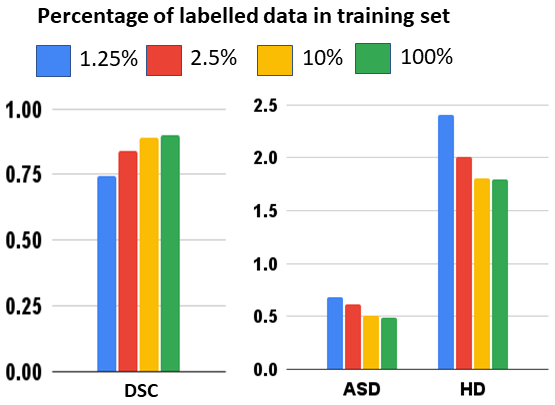}}\hspace{0cm}
    \subfloat[MMWHS]{\includegraphics[scale = 0.415]{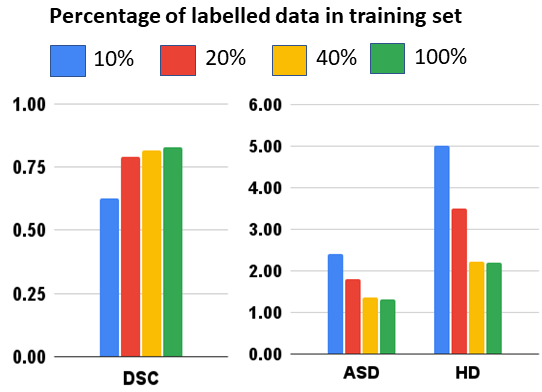}}
    \caption{Quantitative segmentation performance of our proposed method using different percentage of labelled data of ACDC and MMWHS.}
    \label{quant_result_fig}
\end{figure}

To empirically illustrate the effective segmentation ability of the proposed model in a semi-supervised environment, we evaluate our proposed method on the ACDC and MMWHS datasets by training the model using different percentages of labelled data (see \autoref{quant_result_fig}). For the ACDC dataset, the reported DSC, ASD and HD values are 0.746, 0.677, 2.409, respectively while using 1.25$\%$ training samples, 0.842, 0.614, 2.009 corresponding to 2.5$\%$ training labels, 0.889, 0.511, 1.804 for 10$\%$ and 0.902, 0.489, 1.799 while using 100$\%$ samples for training. We consider the 100$\%$ training labels utilization as a fully supervised functioning for comparative analysis. As can be inferred from \autoref{quant_result_fig}\textcolor{blue}{A}, the segmentation performance of the proposed model utilizing only a handful of labels is comparable to the fully supervised counterpart. A similar trend is also observed in \autoref{quant_result_fig}\textcolor{blue}{B}, where the DSC score while using 10$\%$, 20$\%$, 40$\%$ and 100$\%$ training samples are 0.626, 0.791, 0.815 and 0.826 respectively. The reported ASD scores for the respective cases are 2.397, 1.798, 1.355 and 1.317, respectively. The HD score for the 10$\%$ case is 5.001, which is slightly higher, but while using 40$\%$ of the labels for training, the HD score drops down to 2.221, comparable to the fully supervised case with an HD of 2.183.

\subsection{Comparison with State-of-the-art}
To prove the effectiveness of the proposed model, we have performed its comparative analysis with some state-of-the-art models \cite{chen2020simple,hu2021semi,chen2019self,chaitanya2019semi} which is shown in \autoref{comparison_table}. As can be inferred, for the ACDC dataset, the average DSC score as reported by our model while using 1.25$\%$ and 2.5$\%$ training samples is the second best with the state-of-the-art performance given by Global + Local CL \cite{chaitanya2020contrastive}, and PCL \cite{zeng2021positional} respectively. For 10$\%$ training samples, on the other hand, the state-of-the-art performance is reported by the proposed model (DSC=0.889). For the MMWHS dataset, the proposed model surpasses all the existing models in terms of the average DSC score while using 10$\%$, 20$\%$ or 40$\%$ of the training samples, with an average 5$\%$ margin over the second-best performing model. Another important observation is that methods like \cite{zeng2021positional,bai2017semi,hu2021semi} fail to produce satisfactory results using very few annotations (1.25\% and 10\% labelled data for ACDC and MMWHS, respectively). In contrast, our proposed method performs quite consistently as compared to those. We record class-wise DSC and sensitivity to observe the improvements for under-performing classes. Observed DSC and sensitivity are (0.934, 0.883, 0.877) and (0.964, 0.925, 0.911) for classes LV, RV, and MYO respectively using 10\% labelled data of ACDC. Our method achieves an improvement of $\approx2-5\%$ for under-performing classes (RV, MYO) than the baseline and literature. Similar improvements are observed across the classes in other experimental settings, both for ACDC and MMWHS.

\begin{table}[tbp]
\centering
\caption{Comparison of the proposed method with state-of-the-art frameworks
on ACDC and MMWHS datasets.}
\label{comparison_table}
\resizebox{0.9\textwidth}{!}{%
\begin{tabular}{|c|ccc|ccc|}
\hline
\multirow{2}{*}{\textbf{Method}}    & \multicolumn{3}{c|}{\textbf{Average DSC (ACDC)}}                            & \multicolumn{3}{c|}{\textbf{Average DSC (MMWHS)}}                        \\ \cline{2-7} 
                            & \multicolumn{1}{c|}{L=$1.25\%$} & \multicolumn{1}{c|}{L=$2.5\%$} & L=$10\%$ & \multicolumn{1}{c|}{L=$10\%$} & \multicolumn{1}{c|}{L=$20\%$} & L=$40\%$ \\ \hline
Global CL \cite{chen2020simple}                  & \multicolumn{1}{c|}{0.729}   & \multicolumn{1}{c|}{-}      & 0.847 & \multicolumn{1}{c|}{0.5}   & \multicolumn{1}{c|}{0.659} & 0.785 \\ \hline
PCL \cite{zeng2021positional}                        & \multicolumn{1}{c|}{0.671}   & \multicolumn{1}{c|}{\textbf{0.85}}   & 0.885 & \multicolumn{1}{c|}{-}     & \multicolumn{1}{c|}{-}     & -     \\ \hline
Context Restoration \cite{chen2019self}        & \multicolumn{1}{c|}{0.625}   & \multicolumn{1}{c|}{0.714}  & 0.851 & \multicolumn{1}{c|}{0.482} & \multicolumn{1}{c|}{0.654} & 0.783 \\ \hline
Label Efficient \cite{hu2021semi} & \multicolumn{1}{c|}{-}       & \multicolumn{1}{c|}{-}      & -     & \multicolumn{1}{c|}{0.382} & \multicolumn{1}{c|}{0.553} & 0.764 \\ \hline
Data Aug \cite{chaitanya2019semi}                   & \multicolumn{1}{c|}{0.731}   & \multicolumn{1}{c|}{0.786}  & 0.865 & \multicolumn{1}{c|}{0.529} & \multicolumn{1}{c|}{0.661} & 0.785 \\ \hline
Self Train \cite{bai2017semi}                 & \multicolumn{1}{c|}{0.69}   & \multicolumn{1}{c|}{0.749}  & 0.86  & \multicolumn{1}{c|}{0.563} & \multicolumn{1}{c|}{0.691} & 0.801 \\ \hline
Global + Local CL \cite{chaitanya2020contrastive} &
  \multicolumn{1}{c|}{\textbf{0.757}} &
  \multicolumn{1}{c|}{0.826} &
  \multicolumn{1}{c|}{0.886} &
  \multicolumn{1}{c|}{0.617} &
  \multicolumn{1}{c|}{0.710} &
  0.794 \\ \hline
\textbf{Ours} &
  \multicolumn{1}{c|}{{0.746}} &
  \multicolumn{1}{c|}{{0.842}} &
  \textbf{0.889} &
  \multicolumn{1}{c|}{\textbf{0.626}} &
  \multicolumn{1}{c|}{\textbf{0.791}} &
  \textbf{0.815} \\ \hline
\end{tabular}%
}
\end{table}

\subsection{Ablation Studies}
To find out the potency of different components used in the formulation of our proposed scheme, we perform a detailed ablation study, as shown in \autoref{ablation_table}. We use a student-teacher framework as the baseline for all the ablation experiments, which produces an average DSC of $0.819$ and $0.734$ upon evaluation on ACDC and MMWHS datasets, respectively, as a standalone model. Then, we perform two sets of experiments to identify the importance of Robust Class-wise Sampling (RCS) and fuzzy fusion. Instead of fuzzy fusion, as described in \autoref{fuzzy_equation}, first we form the CC score by just simple averaging the three performance indicators: $CC_c = (\mathbb{E}_c+\mathbb{V}_c+\mathbb{C}on_c)/3$. This, along with the RCS, improves the baseline performance by $4.65\%$ and $6.40\%$ in terms of DSC on ACDc and MMWHS, respectively. To fully utilize the potential of RCS, when we evaluate it along with fuzzy fusion, the model outperforms the baseline DSC by a margin of $6.73\%$ and $7.91\%$ respectively on the two datasets. The fuzzy fusion scheme adaptively prioritizes the confidence scores for individual samples rather than using any preset fixed weights to combine the confidence scores. On the other hand, Dynamic Training Stabilization (DTS) and RCS alleviate the problem of class-wise biased training by adaptively prioritizing the under-performing sample classes. This is also justified by the improvements of $\approx 2\%$ and $\approx 3\%$ on ACDC and MMWHS brought by DTS when used on top of RCS with a simple average rule. Furthermore, we present our proposed scheme using DTS and a fuzzy fusion scheme in RCS to achieve the best DSC of $0.889$ and $0.815$ on ACDC and MMWHS, respectively. Additionally, we also observe the variation of model performance by using three confidence indicators individually (refer to supplementary file).          

\begin{table}[tbp]
\centering
\caption{Ablation study to identify the effectiveness of Robust Class-wise Sampling (RCS), Dynamic Training Stabilization (DTS), Fuzzy Fusion on ACDC and MMWHS datasets using 10\% and 40\% labelled data respectively. }
\label{ablation_table}
\resizebox{0.9\textwidth}{!}{%
\begin{tabular}{|c|cc|c|ccc|ccc|}
\hline
\multirow{2}{*}{\textbf{\begin{tabular}[c]{@{}c@{}}Student-Teacher  \\ (Base Model)\end{tabular}}} &
  \multicolumn{2}{c|}{\textbf{RCS}} &
  \multirow{2}{*}{\textbf{DTS}} &
  \multicolumn{3}{c|}{\textbf{ACDC}} &
  \multicolumn{3}{c|}{\textbf{MMWHS}} \\ \cline{2-3} \cline{5-10} 
 &
  \multicolumn{1}{c|}{\textbf{\begin{tabular}[c]{@{}c@{}}Simple \\  Avg. Rule\end{tabular}}} &
  \textbf{\begin{tabular}[c]{@{}c@{}}Fuzzy \\ fusion\end{tabular}} &
   &
  \multicolumn{1}{c|}{\textbf{DSC}} &
  \multicolumn{1}{c|}{\textbf{ASD}} &
  \textbf{HD} &
  \multicolumn{1}{c|}{\textbf{DSC}} &
  \multicolumn{1}{c|}{\textbf{ASD}} &
  \textbf{HD} \\ \hline
\checkmark &
  \multicolumn{1}{c|}{} &
   &
   &
  \multicolumn{1}{c|}{0.817} &
  \multicolumn{1}{c|}{2.697} &
  0.653 &
  \multicolumn{1}{c|}{0.734} &
  \multicolumn{1}{c|}{2.898} &
  2.119 \\ \hline
\checkmark &
  \multicolumn{1}{c|}{\checkmark} &
   &
   &
  \multicolumn{1}{c|}{0.855} &
  \multicolumn{1}{c|}{2.118} &
  0.580 &
  \multicolumn{1}{c|}{0.781} &
  \multicolumn{1}{c|}{2.419} &
  1.662 \\ \hline
\checkmark &
  \multicolumn{1}{c|}{} &
  \checkmark &
   &
  \multicolumn{1}{c|}{0.861} &
  \multicolumn{1}{c|}{1.978} &
  0.541 &
  \multicolumn{1}{c|}{0.792} &
  \multicolumn{1}{c|}{2.328} &
  1.461 \\ \hline
\checkmark &
  \multicolumn{1}{c|}{\checkmark} &
   &
  \checkmark &
  \multicolumn{1}{c|}{0.872} &
  \multicolumn{1}{c|}{1.893} &
  0.522 &
  \multicolumn{1}{c|}{0.809} &
  \multicolumn{1}{c|}{2.291} &
  1.377 \\ \hline

\textbf{\checkmark} &
  \multicolumn{1}{c|}{\textbf{}} &
  \textbf{\checkmark} &
  \textbf{\checkmark} &
  \multicolumn{1}{c|}{\textbf{0.889}} &
  \multicolumn{1}{c|}{\textbf{1.804}} &
  \textbf{0.511} &
  \multicolumn{1}{c|}{\textbf{0.815}} &
  \multicolumn{1}{c|}{\textbf{2.221}} &
  \textbf{1.355} \\ \hline
\end{tabular}%
}
\end{table}

\section{Conclusion}
The scarcity of pixel-level annotations has always been a significant hurdle for medical image segmentation. Besides, a limitation of the deep-learning-based strategies is that they get biased toward the majority class, thereby affecting the overall model performance. To this end, our work addresses both issues by forming a class-wise performance-aware dynamic learning strategy. Experimentation on two publicly available cardiac MRI datasets exhibits the superiority of the proposed method over the state-of-the-art methods. 
In future, we plan to extend the work by designing it as a fine-tuning strategy on top of a contrastive pre-training for more effective utilization of the global context. 

%
\bibliographystyle{splncs04}
\bibliography{paper1371}

\end{document}


\title{Supplementary Material: Addressing Class Imbalance in Semi-supervised Image Segmentation: A Study on Cardiac MRI}
%
%
\author{Hritam Basak\inst{1} \thanks{Corresponding author} \and
Sagnik Ghosal\inst{1} \and
Ram Sarkar\inst{2}}
\authorrunning{H. Basak et al.}
%
\institute{Dept. of Electrical Engineering, Jadavpur University, Kolkata, India \and
Dept. of Computer Science and Engineering, Jadavpur University, Kolkata, India\\
\email{\{hritambasak48, sagnikghosal1999, ramjucse\}@gmail.com}}                                                       
%
\maketitle              
%

\begin{minipage}[c]{0.4\columnwidth}
\centering
\captionof{table}{Variation of DSC w.r.t. $\beta$ in DTS}
\label{eq_13_beta}
\resizebox{\columnwidth}{!}{%
\begin{tabular}{|c|c|c|}
\hline
\textbf{$\beta$} & \textbf{\begin{tabular}[c]{@{}c@{}}ACDC \\ (10\% labelled)\end{tabular}} & \textbf{\begin{tabular}[c]{@{}c@{}}MMWHS\\ (40\% labelled)\end{tabular}} \\ \hline
0.0            & 0.876          & 0.807          \\ \hline
0.5          & 0.882          & 0.810          \\ \hline
1.0            & 0.887          & 0.812          \\ \hline
\textbf{1.5} & \textbf{0.889} & \textbf{0.815} \\ \hline
2.0            & 0.881          & 0.813          \\ \hline
2.5          & 0.879          & 0.808          \\ \hline
3.0          & 0.878          & 0.806          \\ \hline
\end{tabular}%
}
\end{minipage}
\hspace{1cm}
\begin{minipage}[c]{0.4\columnwidth}

\centering
\captionof{table}{Variation of DSC w.r.t. $\lambda$ in RCS}
\label{eq_10_lambda}
\resizebox{\columnwidth}{!}{%
\begin{tabular}{|c|c|c|}
\hline
\textbf{$\lambda$} & \textbf{\begin{tabular}[c]{@{}c@{}}ACDC\\ (10\% labelled)\end{tabular}} & \textbf{\begin{tabular}[c]{@{}c@{}}MMWHS\\ (40\% labelled)\end{tabular}} \\ \hline
1.0            & 0.879          & 0.805          \\ \hline
1.5          & 0.881          & 0.808          \\ \hline
2.0            & 0.886          & 0.811          \\ \hline
\textbf{2.5} & \textbf{0.889} & \textbf{0.815} \\ \hline
3.0            & 0.887          & 0.811          \\ \hline
3.5          & 0.886          & 0.811          \\ \hline
4.0            & 0.884          & 0.810          \\ \hline
\end{tabular}%
}
\end{minipage}
\begin{table}[htbp]
\centering
\caption{Performance for different confidence indicators on Robust Class-wise Sampling using 10\% and 40\% labelled data for ACDC and MMWHS respectively}
\label{performance_ablation}
\resizebox{0.9\textwidth}{!}{%
\begin{tabular}{|c|c|ccc|ccc|}
\hline
\multirow{2}{*}{\textbf{Performance Indicator}} &
  \multirow{2}{*}{\textbf{DTS}} &
  \multicolumn{3}{c|}{\textbf{ACDC}} &
  \multicolumn{3}{c|}{\textbf{MMWHS}} \\ \cline{3-8} 
 &
   &
  \multicolumn{1}{c|}{\textbf{DSC}} &
  \multicolumn{1}{c|}{\textbf{ASD}} &
  \textbf{HD} &
  \multicolumn{1}{c|}{\textbf{DSC}} &
  \multicolumn{1}{c|}{\textbf{ASD}} &
  \textbf{HD} \\ \hline
\multirow{2}{*}{Entropy ($\mathbf{E}$)} &
   &
  \multicolumn{1}{c|}{0.847} &
  \multicolumn{1}{c|}{1.836} &
  0.539 &
  \multicolumn{1}{c|}{0.788} &
  \multicolumn{1}{c|}{2.246} &
  1.474 \\ \cline{2-8} 
 &
  $\checkmark$ &
  \multicolumn{1}{c|}{0.866} &
  \multicolumn{1}{c|}{1.821} &
  0.529 &
  \multicolumn{1}{c|}{0.793} &
  \multicolumn{1}{c|}{2.234} &
  1.368 \\ \hline
\multirow{2}{*}{Variance ($\mathbf{V}$)} &
   &
  \multicolumn{1}{c|}{0.855} &
  \multicolumn{1}{c|}{1.837} &
  0.540 &
  \multicolumn{1}{c|}{0.776} &
  \multicolumn{1}{c|}{2.244} &
  1.477 \\ \cline{2-8} 
 &
  $\checkmark$ &
  \multicolumn{1}{c|}{0.864} &
  \multicolumn{1}{c|}{1.826} &
  0.531 &
  \multicolumn{1}{c|}{0.787} &
  \multicolumn{1}{c|}{2.237} &
  1.372 \\ \hline
\multirow{2}{*}{Confidence ($\mathbf{C}on$)} &
   &
  \multicolumn{1}{c|}{0.866} &
  \multicolumn{1}{c|}{1.825} &
  0.527 &
  \multicolumn{1}{c|}{0.784} &
  \multicolumn{1}{c|}{2.327} &
  1.441 \\ \cline{2-8} 
 &
  $\checkmark$ &
  \multicolumn{1}{c|}{0.878} &
  \multicolumn{1}{c|}{1.811} &
  0.519 &
  \multicolumn{1}{c|}{0.809} &
  \multicolumn{1}{c|}{2.228} &
  1.361 \\ \hline
\multirow{2}{*}{\textbf{Cumulative Confidence ($CC$)}} &
   &
  \multicolumn{1}{c|}{0.879} &
  \multicolumn{1}{c|}{1.809} &
  0.516 &
  \multicolumn{1}{c|}{0.810} &
  \multicolumn{1}{c|}{2.230} &
  1.362 \\ \cline{2-8} 
 &
  \textbf{$\checkmark$} &
  \multicolumn{1}{c|}{\textbf{0.889}} &
  \multicolumn{1}{c|}{\textbf{1.804}} &
  \textbf{0.511} &
  \multicolumn{1}{c|}{\textbf{0.815}} &
  \multicolumn{1}{c|}{\textbf{2.221}} &
  \textbf{1.355} \\ \hline
\end{tabular}%
}
\end{table}

\begin{table}[htbp]
\centering
\caption{Variation of model performance on ACDC and MMWHS datasets while using different fuzzy functions in Robust Class-wise Sampling}
\label{fuzzy_comparison}
\resizebox{0.9\textwidth}{!}{%
\begin{tabular}{|c|ccc|ccc|}
\hline
\multirow{2}{*}{\textbf{Fuzzy   Function}} &
  \multicolumn{3}{c|}{\textbf{Average DSC (ACDC)}} &
  \multicolumn{3}{c|}{\textbf{Average DSC (MMWHS)}} \\ \cline{2-7} 
 &
  \multicolumn{1}{c|}{\textbf{L=1.25\%}} &
  \multicolumn{1}{c|}{\textbf{L=2.5\%}} &
  \textbf{L=10\%} &
  \multicolumn{1}{c|}{\textbf{L=10\%}} &
  \multicolumn{1}{c|}{\textbf{L=20\%}} &
  \textbf{L=40\%} \\ \hline
Mitscherlich \cite{b4} &
  \multicolumn{1}{c|}{0.737} &
  \multicolumn{1}{c|}{0.834} &
  0.881 &
  \multicolumn{1}{c|}{0.612} &
  \multicolumn{1}{c|}{0.785} &
  0.798 \\ \hline
Blumberg \cite{b2} &
  \multicolumn{1}{c|}{0.740} &
  \multicolumn{1}{c|}{0.839} &
  0.886 &
  \multicolumn{1}{c|}{0.618} &
  \multicolumn{1}{c|}{0.789} &
  0.812 \\ \hline
Weibull \cite{b5} &
  \multicolumn{1}{c|}{0.739} &
  \multicolumn{1}{c|}{0.837} &
  0.885 &
  \multicolumn{1}{c|}{0.618} &
  \multicolumn{1}{c|}{0.788} &
  0.813 \\ \hline
\textbf{Gompertz} &
  \multicolumn{1}{c|}{\textbf{0.746}} &
  \multicolumn{1}{c|}{\textbf{0.842}} &
  \textbf{0.889} &
  \multicolumn{1}{c|}{\textbf{0.626}} &
  \multicolumn{1}{c|}{\textbf{0.791}} &
  \textbf{0.815} \\ \hline
\end{tabular}%
}
\end{table}

\begin{figure}[!h]
    \centering
    \includegraphics[width=0.9\columnwidth]{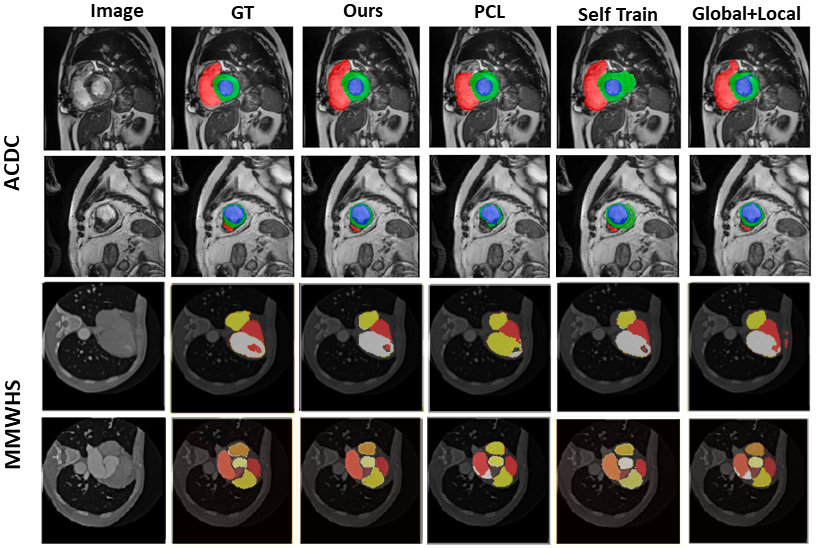}
    \caption{Visual comparison of our proposed method with the available ground truth (GT), along with several other state-of-the-art methods: PCL \cite{b6}, Self Train \cite{b1}, and Global + Local CL \cite{b3}}
    \label{comparison_figure}
\end{figure}

\bibliographystyle{splncs04}
%